# Evaluating Single and Multi-Omics Based Explainable Artificial Intelligence (MOXAI) for Molecular Subclass Classification of Adult-Type Diffuse Gliomas

**Md Zahangir Alom, Quynh T. Tran, Breuer Alexandar, and Brent A. Orr***
Department of Pathology, St. Jude Children's Research Hospital, Memphis, TN 38105 USA.
Email: Zahangir.Alom@stjdue.org, Quynh.Tran@stdjue.org, Alex.Breuer@stjude.org, and Brent.Orr@stjude.org

*Abstract*— **DNA methylation (DNAM) profiling has emerged as a powerful diagnostic tool for classifying brain and solid tumors. However, existing computational models typically analyze methylation and copy number variation (CNV) data separately, failing to capture the complementary information their integration could provide. Moreover, current classification models lack mechanisms for within-class risk assessment analogous to traditional tumor grading, and no established explainability method can attribute classification decisions to specific genomic loci. In this paper, we present MOXAI (Multi-Omics Based Explainable AI), a deep learning framework that integrates DNA methylation and copy number data from methylation arrays to classify molecular subtypes of adult-type diffuse gliomas, alongside single-modality variants for comparison. Using a cohort from The Cancer Genome Atlas (TCGA), we trained ResNet50, DINOv2, and Graph Attention Network (GAT) models on methylation data alone, copy number data alone, and combined multimodal data. We further developed explainable AI (XAI) methods based on class activation maps (CAMs) and gradient-weighted CAM (Grad-CAM) to identify the specific CpG sites, genes, and chromosomal regions most relevant to each classification decision. The multimodal model achieved up to 92.98% cross-validation accuracy, outperforming models trained on CNV data alone. DINOv2 showed the strongest generalization, reaching 94.25% accuracy (confidence >0.9) on independent validation sets. XAI results aligned with established molecular features of adult-type diffuse glioma subtypes, confirming the biological interpretability of the framework.**

**Keywords—*brain tumor classification, DNA methylation, copy number variation, multi-omics, explainable AI, DiNOv2, ResNet50, GAT, CAM and Grad-CAM.***

## I. Introduction

Brain tumors represent a diverse group of malignancies whose accurate classification is critical for determining prognosis and guiding therapeutic decisions. In recent years, DNA methylation (DNAM) profiling has become an essential foundation of brain tumor diagnostics, as evidenced by its central role in the 2021 WHO classification of central nervous system (CNS) tumors [1]. DNA methylation arrays, particularly the Illumina 450K and EPIC platforms, capture genome-wide methylation patterns with hundreds of thousands of CpG sites, providing a rich molecular fingerprint that is highly discriminative across tumor types and subtypes [2]. Although this diagnostic power exists, the present workflows show several limitations. First, while both methylation and copy number variation (CNV) data are routinely derived from the same DNA methylation array experiment, CNV results are typically reported independently of classification scores without systematic integration. Second, selected molecular tumor classes are known to carry differential clinical risk, yet existing models do not output within-class risk assessments analogous to histological grading. Third, methods to quantify the relative contribution of specific CpG sites or chromosomal regions to a classification decision—a critical requirement for clinical interpretability—have not been established. Deep convolutional neural networks (DCNNs) have demonstrated strong performance across a range of medical imaging and genomics tasks [3, 13, 16]. However, applying them to multi-omics integration for tumor classification—while maintaining meaningful explainability—remains an open research challenge. Current explainable AI (XAI) methods for omics data generally lack the spatial resolution required to attribute model predictions to individual genomic loci [4]. A central obstacle is scale: the DNA methylation (DNAM) dataset used here contains 408,768 probes, which makes conventional sequential DNNs computationally prohibitive due to their enormous parameter counts and GPU memory demands. This high dimensionality also makes applying explainable methods to such architectures particularly difficult. To overcome these issues, we developed MOXAI (Multi-Omics based eXplainable Artificial Intelligence), which transforms genomic features into a 2D representation. This transformation allows us to leverage established image-based deep learning architectures and XAI techniques—such as class activation maps (CAM) and Grad-CAM—while retaining a large proportion of the available probes.

Our approach makes three main contributions: (1) a unified representation that enables joint modeling of DNA methylation and copy number data through both single- and multi-modal systems; (2) an evaluation of ResNet50, the DINOv2 transformer, and a Graph Attention Network (GAT) for diffuse glioma classification using individual and combined modalities; and (3) an XAI framework based on CAM and Grad-CAM [15] that localizes the most diagnostically informative CpG sites and chromosomal regions for each tumor class. We evaluate this approach on molecular subtypes of adult-type diffuse gliomas from The Cancer Genome Atlas (TCGA), with validation on independent glioma datasets, showing that multimodal integration produces better interpretable, and accurate classification. Notably, the explanations generated by our system align closely with established molecular features of adult glioma subtypes.

## II. Background and Related Work

### A. DNA Methylation-Based Brain Tumor Classification

The initial Heidelberg classifier, introduced by Capper et al. [2,16], demonstrated that genome wide methylation profiling can reliably classify brain tumors into over 91 distinct methylation classes with high accuracy. This approach, now widely adopted in neuropathology practice, uses a random forest classifier trained on beta-values from Illumina methylation arrays. While this system has transformed tumor diagnostics, it does not incorporate copy number information into the classification score and does not provide locus-level explainability.

### B. Copy Number Variation in Brain Tumors(Diffuse gliomas)

CNV data derived from methylation arrays capture large-scale chromosomal alterations, including gains, losses, and amplifications, which are hallmarks of specific tumor subtypes. For instance, IDH-mutant astrocytoma's frequently demonstrate loss of chromosomes 9p and 10q, while IDH-mutant oligodendrogliomas are characterized by 1p/19q co-deletion [5]. Another example of glioblastoma subtypes demonstrates distinct patterns, including gain of chromosome 7 which includes the EGFR locus and whole chromosome loss of chromosome 10 including the PTEN locus. Although this has diagnostic relevance, CNV data have not been systematically integrated into deep learning classification pipelines alongside methylation data.

### C. Multi-Omics Integration with Deep Learning

Multi-omics integration is a way of simultaneously analyzing multiple molecular data, which has been explored in various cancer genomics contexts in recent years [6, 14]. Approaches include early fusion (concatenation of feature vectors), late fusion (ensemble of modality-specific models), and intermediate fusion (joint representation learning). Deep learning architectures, particularly CNNs and attention-based models, have shown promise in learning shared representations from heterogeneous omics data. However, principled multi-omics integration for methylation array data, exploiting the spatial ordering of genomic probes, has not been previously demonstrated.

### D. Explainable AI for Genomics

XAI methods for omics data have been reviewed by Toussaint et al. [4], who identified gradient-based attribution, attention mechanisms, and feature importance methods as predominant approaches. We used class activation mapping (CAM) [7], and its variants Grad-CAM provide spatially resolved attribution maps that have been extensively applied to image classification tasks [15]. Adapting these methods to genome-ordered probe matrices enables the identification of specific genomic regions driving classification decisions a form of explainability that aligns naturally with chromosomal biology.

## III. Methods

### A. Dataset

This study used 551 adult-type diffuse glioma samples from The Cancer Genome Atlas (TCGA), spanning six methylation classes shown in Fig. 1: astrocytoma IDH-mutant (AIDH), astrocytoma IDH-mutant high grade (AIDH-HG), oligodendroglioma IDH-mutant (OIDH), and three glioblastoma subtypes—mesenchymal (GBM-MES), receptor tyrosine kinase I (GBM-RTKI), and receptor tyrosine kinase II (GBM-RTKII). Samples were profiled using the Illumina HumanMethylation450 BeadChip array. Following standard preprocessing protocols, sex chromosome probes and known mis-hybridizing probes were excluded, retaining 408,768 probes with natural alignment to somatic chromosomes across samples. A key contribution of this work is a unified matrix representation that enables convolutional architectures to jointly process DNA methylation and copy number variation (CNV) data. Probes were ordered by genomic location (chromosome coordinate) and arranged into a 640×640-pixel matrix, producing a spatially coherent genomic image in which proximal pixels correspond to genomically adjacent loci. DNA methylation was normalized as beta-values (0 to 1, from unmethylated to fully methylated), while copy number was encoded as the normalized segment value at each corresponding CpG site, derived via standard segmentation algorithms applied to array intensity data [8]. Since the number of probes does not perfectly fill a square matrix, residual empty positions were zero-padded. Single-modality experiments used single-channel matrices (1×640×640), while multimodal experiments combined DNAM and CNV into two-channel matrices (2×640×640). Model performance was validated on independent test sets from [19], comprising 492 samples across three glioblastoma molecular subtypes: GBM-MES (158), GBM-RTKI (120), and GBM-RTKII (214).

### B. Model Architecture

We employed three well-established deep learning architectures in this study: ResNet50 [9], the DINOv2 transformer [11], and Graph Attention Networks (GAT) [18]. Each model was adapted for single- and dual-channel input by modifying its first convolutional layer accordingly, with weights randomly initialized and trained from scratch for both the single-channel DNAM model and the two-channel multimodal configuration. ResNet50 consists of 50 layers organized into residual blocks, which use skip connections to mitigate vanishing gradients while enabling the network to learn hierarchical genomic features [9,11]. This architecture was chosen in part for its ability to handle large-scale inputs while addressing gradient vanishing issues [9,15]. The final fully connected layer was configured to output logits for the six glioma classes. Models were trained using cross-entropy loss with L2 regularization and optimized with Adam [12], starting at a learning rate of $1\times10^{-4}$ and decaying by a factor of 0.1 every 20 epochs. Training ran for 150 epochs with a batch size of 8. For the GAT model [18], probes mapped to null genes were excluded, reducing the probe set from 408,768 to 297,730. These remaining probes were organized hierarchically: first aggregated into gene-level representations across 17,630 unique genes (from a 20K gene embedding), then mapped to their corresponding chromosomes (22 in total). A transformer encoder processed these 22 chromosome-level "tokens," and an attention pooling layer condensed its outputs into a single vector, which was then classified using an MLP classifier.

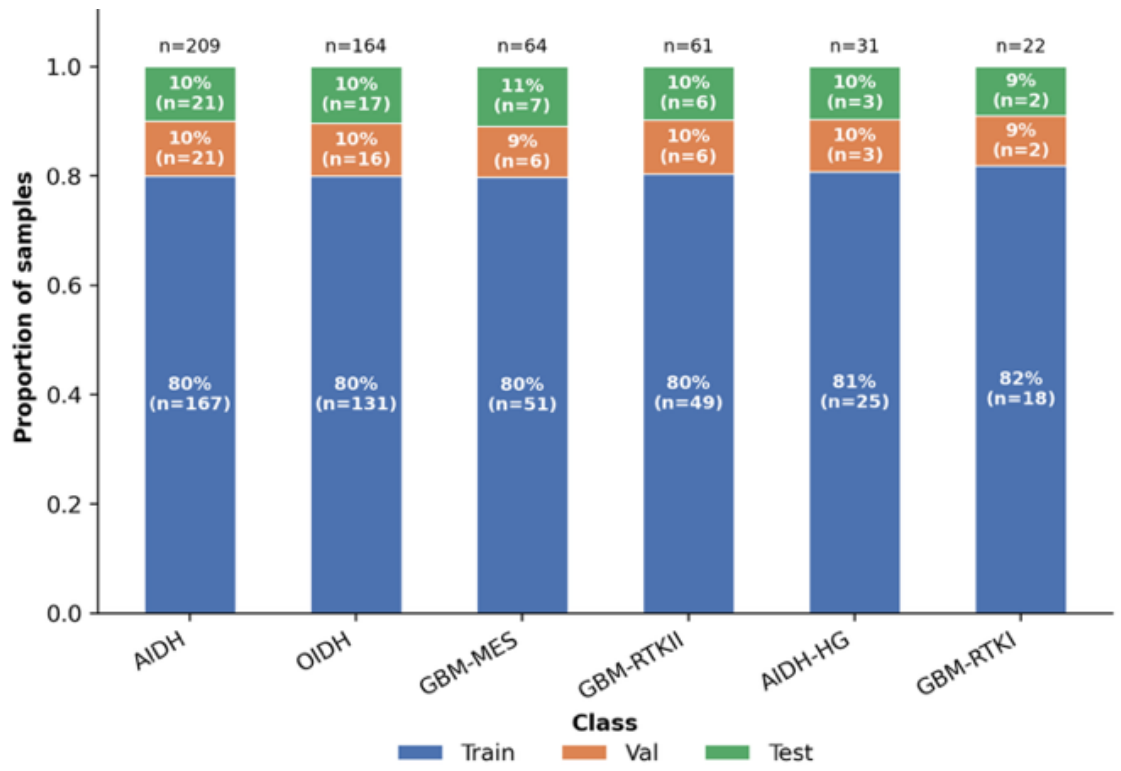

Fig. 1. Data statistics for Train/validation/test ratio per class: The total number of samples of 551, common samples were selected based on the availability of both CNV and DNAM data.

### C. Experimental Design

The dataset was split 80:20 for training and testing with stratification by class. Training was performed with 10-fold cross-validation; both best accuracy and mean accuracy (±standard deviation) across folds are reported. Three experimental conditions were evaluated: (1) CNV-only single-channel model, (2) DNAM-only single-channel model, and (3) combined CNV+DNAM two-channel model. Performance was assessed using accuracy, precision, recall, and F1 score on the held-out test set. All experiments were conducted using PyTorch 2.8 on NVIDIA A100 GPUs.

### D. Explainable AI Methods

To interpret model predictions, we applied class activation mapping (CAM) to the feature representations extracted from the bottleneck layer of the ResNet50 architectures (the 16×16×1024 activation volume preceding the global average pooling layer) and DiNOv2. For each class, the average activation across the spatial dimensions was computed and mapped back to the original 640×640 input space using bilinear up sampling, producing a heatmap of importance scores for each CpG site. Importance score heatmaps ranged from low (blue) to high (red) importance. For each modality and each class, binary masks were generated by applying a threshold of 0.85 (empirically selected) to the normalized importance map, isolating the most diagnostically informative genomic regions. The chromosomal distribution of high-importance probes was then analyzed, and chromosome-specific CAM distributions were plotted to identify the most relevant chromosomal regions for each glioma subtype. For OIDH, which is characterized by 1p/19q co-deletion, chromosome-arm-level resolution was achieved by computing binned average CAM values across chromosomes 1 (39,743 CpGs, binned at 1,000 probes) and 19 (22,920 CpGs, binned at 500 probes).

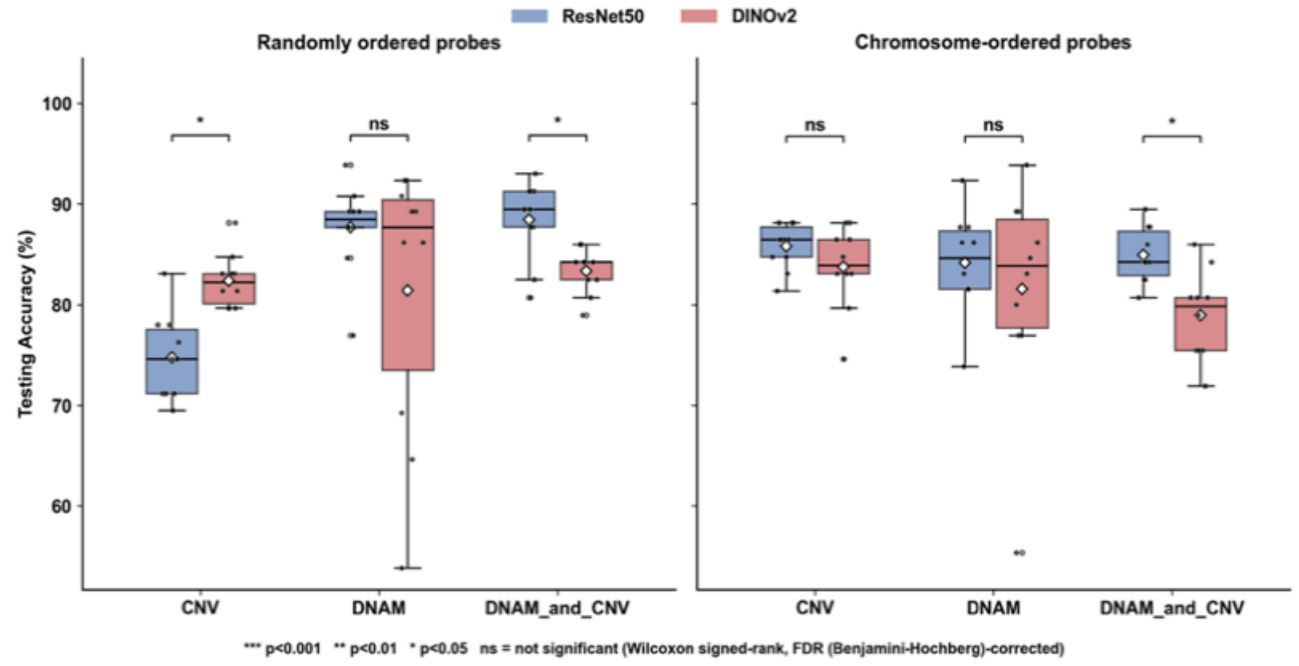

Fig. 2. Validation testing performance comparison ResNet50 vs DiNov2: demonstrates the accuracies for both random ordered probes and probes ordered by chromosome and mapping inform.

## IV. Results

### A. Classification Performance

Fig. 2 summarizes the classification accuracy of the three ResNet50 and DiNOv2 models under both data representation strategies (most significant probes ordered vs. probes in chromosome order) across 10-fold cross-validation and holdout test prediction. The DNAM-only model achieved the best accuracy across both representation strategies (93.84% for the differentially significant probes approach and 92.30% for chromosome-ordered probes), the DiNOv2 shows 93.86%. The multimodal ResNet50 models demonstrated the average performance 88.42% random ordered and 84.91% chromosome ordered, respectively) consistently outperforming the CNV-only baseline (mean of 74.74% and 85.76%), whereas DiNOv2 shows 83.33% and 78.94%. These results confirm that methylation information is the dominant signal for glioma subtype discrimination, while CNV data contributes complementary information that stabilizes performance under the chromosome-ordered representation, achieving lower standard deviation (±2.62 vs. ±4.76 for DNAM alone). The CNV-alone ResNet50 model showed the highest performance gap between representation strategies, with chromosome-ordered probes (mean 85.76%) outperforming significance-ordered probes (average of 74.74%) whereas the DiNOv2 shows an average performance of 82.37% for random ordered probes \and 83.72% for chromosome ordered. We observed statistical significance (p-value <0.05) performance for ResNet50 model with CNV alone and multimodal system, and DiNOv2 based multimodal model. The GAT model showed the testing accuracy of 62.6 ± 2.27 and 75.57 ± 1.27 for DNAM alone and multi-modal models. The outcomes suggest that spatial genomic context—preserved in chromosome-ordered representation—is particularly informative for copy number-based classification, consistent with the regional nature of chromosomal gains and losses.

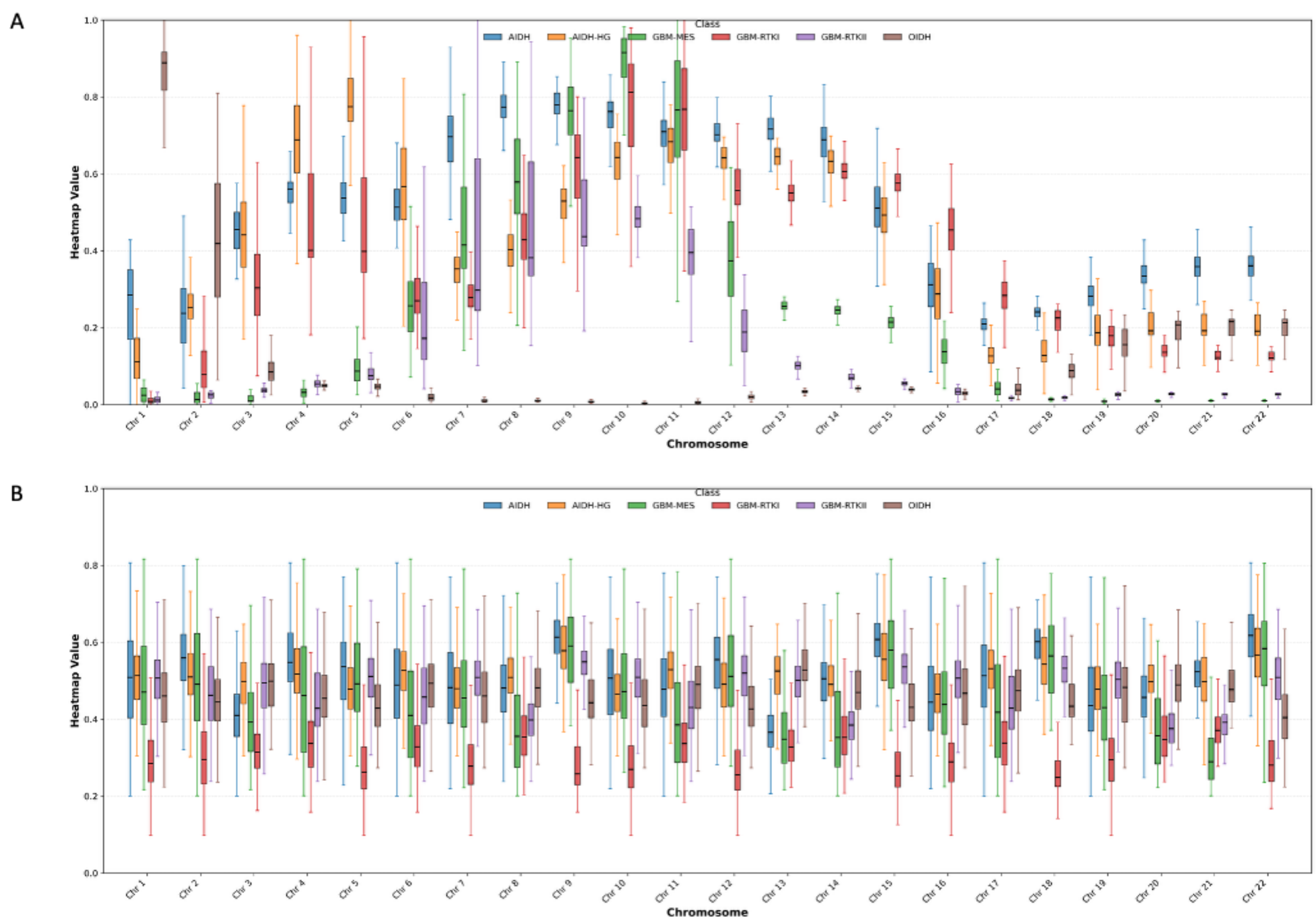

Fig. 3. The chromosome-specific class activation map (CAM) for ResNet50 and DiNOv2 models: (A) shows the class specific box plot for CAM values generated using ResNet50 model for six classes. (B) demonstrates the class specific CAM values generated from DiNOv2. The x-axis represents the chromosome IDs, and the y-axis represents the CAMs value.

*B. XAI Analysis: CNV Model*

Fig. 3 shows the chromosome-specific distribution of CAM importance values derived from the ResNet50-CNV and DINOv2 models, offering insight into which chromosomal regions most strongly drive classification decisions for each glioma subtype. As shown in Fig. 3(A), AIDH (blue) and AIDH-HG (orange) show elevated CAM values concentrated in the mid-range chromosomes (~7–13), suggesting these regions carry substantial discriminative signal for IDH-mutant subtypes. Specifically, chromosome 9 signal is most elevated for AIDH tumors and AIDH-HG tumors but nearly absent for OIDH tumors. This is consistent with the known role of *CDKN2A* deletion on chromosome 9 in discriminating high-grade behavior in IDH-mutant astrocytomas. A similar role has not been established for OIDH tumors. DINOv2, by contrast shows significantly less separation of importance scores compared to the ResNet50 based models and highlights a different set of chromosomes including 9, 15, 18, and 22 as most important. OIDH (gray) displays a markedly different pattern in both models, with CAM ResNet50 values peaking sharply at chromosome 1 before declining and remaining low across most other chromosomes, indicating a more localized region of importance. The importance of chromosome 1 is not clearly observed in the DINOv2 model. As 1p and 19q deletion are hallmarks of OIDH identity, this suggests the ResNet50 based model more faithfully discriminates importance in this tumor type. The glioblastoma subtypes also show distinct chromosomal signatures between models: Astrocytoma subtypes show a pronounced peak centered around chromosomes 7–11, with a peak at chromosome 10. This aligns with the established importance of chromosome 10 loss in GBM, IDH-wiltype subtypes. GBM-RTKI (red) shows a broader, more distributed elevation spanning chromosomes 3–14. Notably both chromosomes 4 and 9 show increased signal, consistent with the enrichment for *PDGFRA* amplification and *CDKN2A* deletion in that subtype. DINOv2 assigned overall lower importance to GBM-RTKI but notably showed its highest value at chromosome 8. Together, these class-specific CAM distributions suggest that the ResNet50 based CAM model learns distinct chromosomal patterns of copy number variation for each glioma subtype and is superior in discriminating regions of importance compared to the DINOv2 CAM models. In addition to CAM, we also evaluateReNet50 based Grad-CAM outputs from the CNV models, and exhibited some notable variability between the two. Fig. 4(A). For AIDH and AIDH-HG, CAM and Grad-CAM outputs, demonstrating near complements of importance regions. For OIDH, the highest importance for Grad-CAM was assigned to chromosome 1, while additionally emphasized chromosomes 9, 10, and 11; notably, the canonical 1p/19q co-deletion signature of oligodendroglioma was also detectable at chromosome 19 using the Grad-CAM based importance. Among glioblastoma subtypes, the CNV model identified chromosome 7 as most important for GBM-RTKI and GBM-RTKII, consistent with EGFR amplification enriched in the RTKII subtype. GBM-MES showed high importance across chromosomes 4, 5, 10, and 11, whereas ResNet50-Grad-CAM emphasized chromosome 7 specifically, a finding notably absent from the CAM based outpug. These findings indicate that the CNV model learned biologically meaningful chromosomal signatures despite end-to-end training without explicit chromosomal annotation.

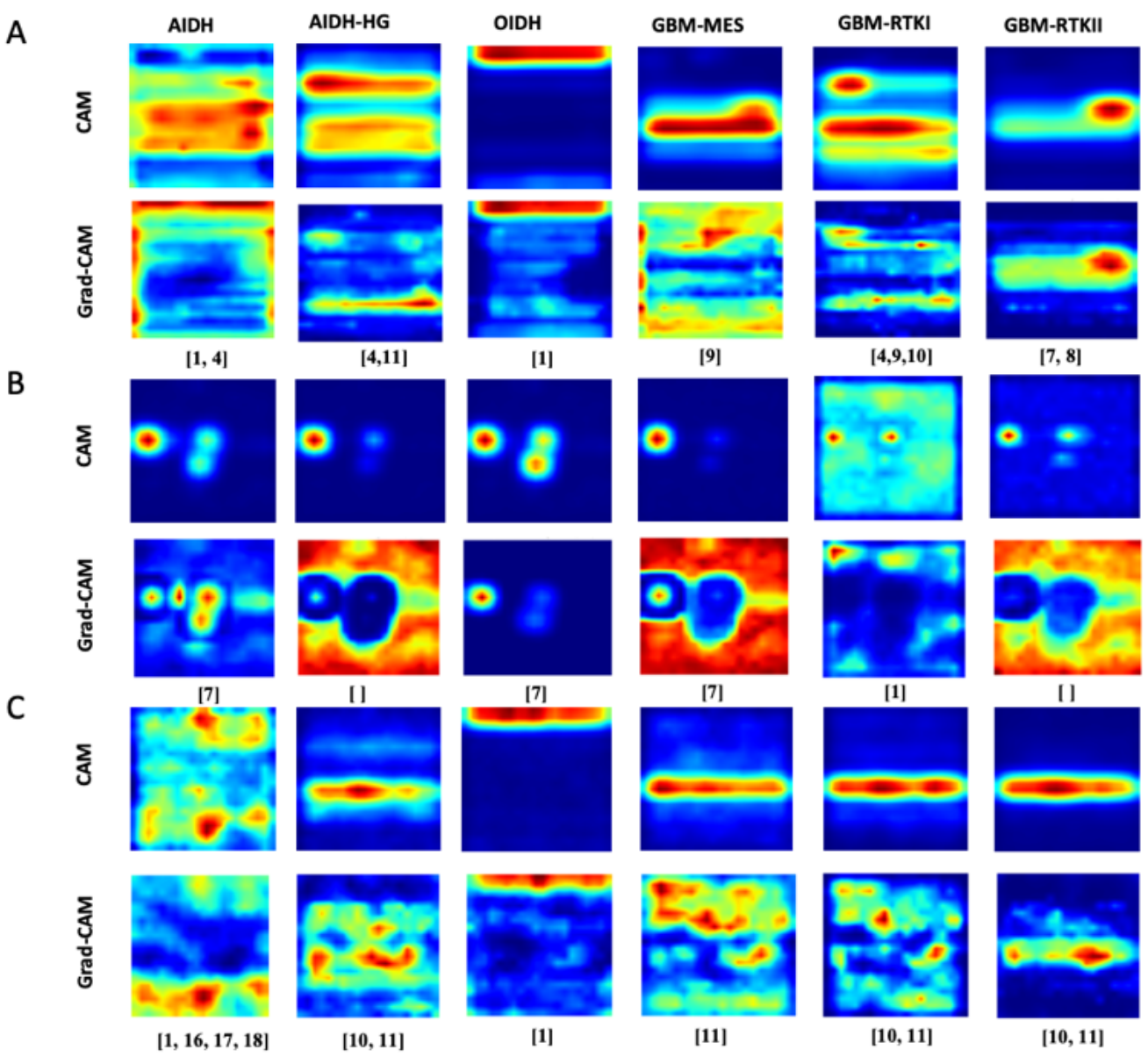

Fig. 4. The XAI results visual comparison between CAM and Grad-CAM methods heatmaps generated from the ResNet50 models. The heatmap values range from blue (low importance) to red (high importance). For each modality, masks were generated for the topmost important site by applying a threshold of 0.85. The most significant common chromosome IDs are shown at the bottom of the respective masks: (A) XAI results from the CNV alone, (B) represents the XAI results from the DNAM alone, and (C) the results from the combined CNV and DNAM modalities.

Fig. 4 presents the chromosome-specific distribution of CAM and Grad-CAM importance derived from the bottleneck-layer representations of the ResNet50 (A) CNV-alone, (B) DNAM-alone, and (C) CNV+DNAM models, illustrating which chromosomal regions most strongly influence classification for each subtype. Overall, these class-specific CAM and Grad-CAM patterns align with the known biological heterogeneity across glioma classes, supporting the interpretability of the CNV-based deep learning model in identifying clinically relevant genomic regions.

### *C. XAI Analysis: DNAM Model*

The DNAM-based XAI analysis with ResNet50-CAM revealed that chromosome 7 emerged as the dominant importance region across multiple GBM subtypes (GBM-MES, GBM-RTKI, GBM-RTKII, and OIDH), reflecting the central role of EGFR locus methylation and associated regulatory elements in glioma biology which is shown in Fig. 4(B). For AIDH and AIDH-HG, chromosomes 7 and 8 showed highest activation. The convergence of multiple subtypes on chromosome 7 in the methylation model is consistent with the frequency of epigenetic dysregulation at this locus across glioma classes. The Grad-CAM outputs showed consistency with the CAM output in some subtypes, but the AIDH-HG, GBM-RTKII, and GBM-RTKI, and GBM-RTKII outputs from using GRAD-CAM were diffuse in comparison to the more restricted profiles from CAM. This may reflect the importance of copy number abnormalities to define these subtypes therefore relying on more distributed DNAM signal to infer class in the absense of CNV data.

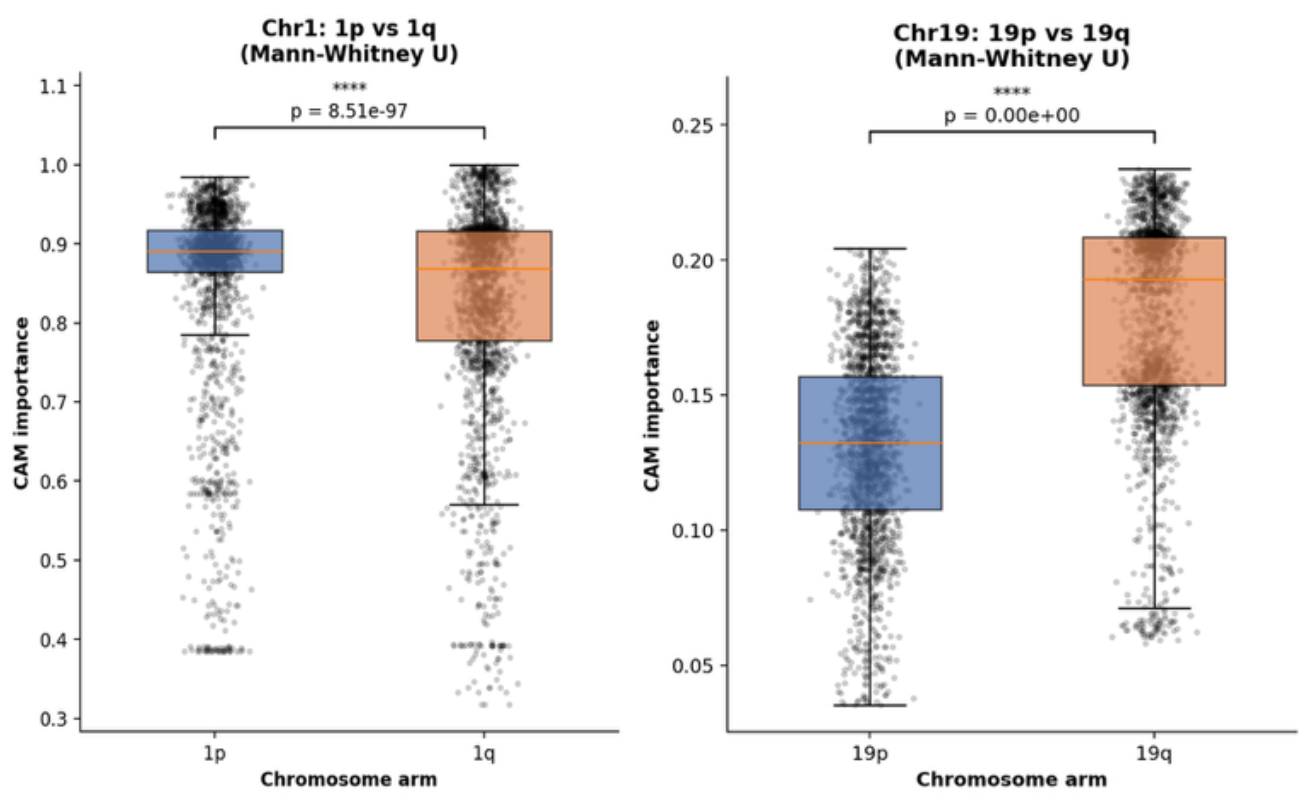

Fig. 5. Binned CpGs-specific ResNet50 average activation map (CAM) values for chromosome 1p vs 1q and 19p vs 19q to determine the chromosome arm-specific activation for OIDH from Fig 3(A)-OIDH. A) Statistical significance analysis with Mann-Whitney method for chromosome 1, (B) Chromosome 19. The x-axis represents the CpGs IDs for average representation, and the y-axis shows the CAM values, respectively.

### *D. XAI Analysis: Multimodal Model*

The combined CNV+DNAM model produced XAI heatmaps that integrated chromosomal and methylation signals from both modalities using CAM and Grad-CAM, yielding importance maps with greater class specificity. The most important chromosomal regions identified in the multimodal masks were: AIDH (chromosomes 1, 16, 17, 18), AIDH-HG (chromosomes 10, 11), OIDH (chromosome 1), GBM-MES (chromosomes 10, 11), GBM-RTKI (chromosomes 10, 11), and GBM-RTKII (chromosomes 7, 10, 11) in Fig. 4(C). The importance of chromosomes 10 and 11 across multiple GBM subtypes in the multimodal model reflects PTEN loss (10q) and CDK6/CCND1 amplification (11q), both of which are relevant to GBM pathobiology. Chromosome 1 signal was maintained for OIDH in the multimodal model—highlighting the importance of copy number abnormalities to define class in this tumor type. . Interesting, high signal regions emerged in the multimodal models that were not reflected in the individual CNV or DNAM models. This was most evident in the AIDH model in which chromosome 18 exhibited elevated importance in the combined model. Although chromosome 18 is not known to play a role in AIDH tumors but holds potential for downstream interrogation.

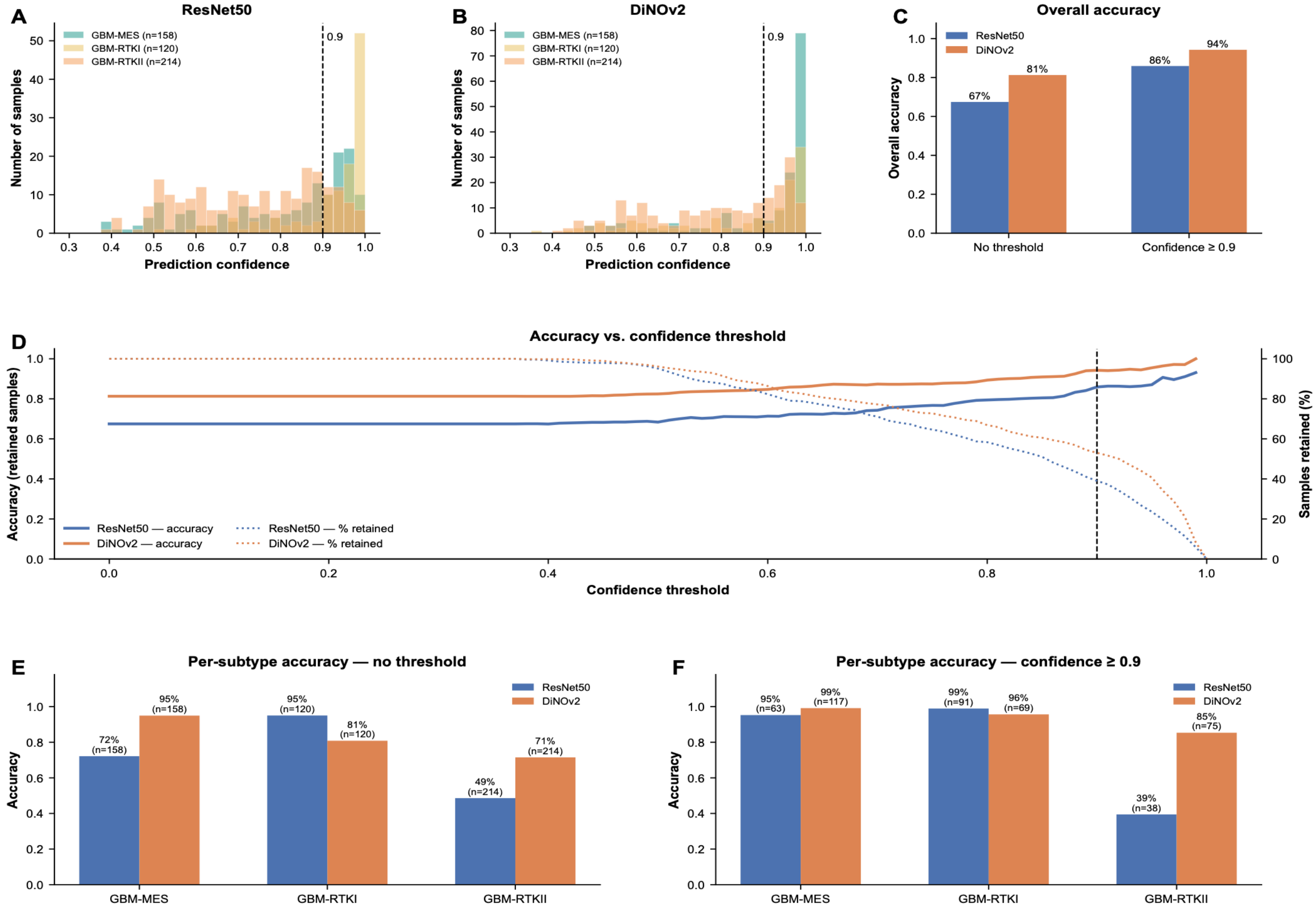

**Fig. 6. Testing performance comparison ResNet50 vs DiNov2 for independent datasets**: (A) and (B) represents the distribution of prediction confidence. (C) demonstrates the accuracies for no threshold and confidence >0.9. (D) Accuracy vs confidences. Per class accuracy no threshold and with threshold are shown in (E) and (F), respectively.

## E. *Chromosome-Arm Resolution for OIDH*

To further characterize the 1p/19q signature in OIDH [17], we computed statistical significance with Mann-Whitney method using the ResNet50-CAM importance values at chromosome-arm resolution that is shown in Fig. 5. For chromosome 1 (39,743 probes, from 1 to 23940 probes for 1p and the remaining for 1q), we observed elevated activation at the p-arm (proximal bins), consistent with the telomeric 1p deletion characteristic of oligodendrogliomas. For chromosome 19 (22,920 probes, from 0 to 11212 probes for 19p and the remaining for 19q), elevated activation was observed across both arms, with the q-arm showing a distinct activation pattern with statistical significance (p-value < 0.05).

## F. *Independent Validation*

We tested the best-performing ResNet50, DINOv2, and GAT models on an independent validation dataset comprising 492 glioblastoma molecular subtype samples, with results reported in Fig. 6. The distribution of prediction confidence for glioblastoma subtypes, shown in Fig. 6(A), indicates that DINOv2 classifies correctly with higher confidence than the other models. Class-specific accuracies are also presented in Fig. 6(A). Overall, ResNet50 achieved a testing accuracy of 67.48%, rising to 85.94% when restricted to high-confidence predictions (confidence > 0.9). DINOv2 outperformed this, achieving an overall accuracy of 81.30% and a high-confidence accuracy of 94.25% (Fig. 6(C)). Fig. 6(D) illustrates the relationship between accuracy, confidence threshold, and sample coverage, while class-specific accuracies are shown in Fig. 6(E, F). The GAT model performed weakest on the independent validation set, with an overall testing accuracy of 62%.

## V. Discussion

MOXAI represents a meaningful advance in the computational analysis of DNA methylation array data for brain tumor diagnostics. By encoding both methylation and CNV data as spatially coherent genomic images and applying ResNet50, DINOv2, and GAT classifiers, we demonstrate that multi-omics integration is both feasible and beneficial within this data modality. ResNet50 achieved the highest overall accuracy, while DINOv2 showed better generalization, yielding the strongest performance on independent test samples. Notably, the multimodal model achieved more stable performance—reflected in lower standard deviation across cross-validation folds—than the methylation-only model, suggesting that CNV data provides complementary regularization even when it does not consistently boost peak accuracy. This stability is clinically meaningful, since diagnostic models must perform reliably across heterogeneous patient populations. The framework stands out for several reasons. First, chromosome-ordered probe matrices allow convolutional layers to learn features that reflect the genome's spatial organization—a biologically motivated inductive bias that we show improves CNV-based classification. Because copy number alterations manifest as contiguous chromosomal segments, their spatial locality is naturally captured by convolutional operations over genomically ordered matrices. Second, our XAI analysis shows that MOXAI learns interpretable, biologically meaningful features: the chromosomal regions identified as most important for each glioma subtype align closely with known molecular hallmarks, including EGFR gain/amplification (chr7) in GBM subtypes, PTEN loss (chr10q) across high-grade gliomas, and 1p/19q co-deletion in OIDH. This concordance validates the model's biological fidelity and supports the potential clinical utility of XAI outputs in molecular diagnostic workflows or studies of tumor biology. Overall, ResNet50-based models proved more interpretable than DINOv2-based models. One likely explanation lies in architectural differences: ResNet50 uses relatively small 3×3 convolutional kernels throughout the network, enabling more localized feature representation. DINOv2, by contrast, computes importance over 16×16 patches, yielding substantially coarser spatial resolution. This coarser granularity makes DINOv2 roughly five times more likely to assign importance across grid boundaries, causing features from adjacent chromosomal regions to merge or "collapse"—which likely explains its lower interpretability in this study.

This work focuses on adult-type diffuse gliomas as a proof-of-concept, but the MOXAI framework is readily generalizable to other tumor types profiled by methylation arrays. Pairing MOXAI with recently developed CNV-calling software [8] could further streamline diagnostic workflows. Several limitations warrant acknowledgment. The TCGA dataset, while well-characterized, consists predominantly of adult samples and may not fully capture the diversity of tumors encountered clinically. Additionally, class imbalance—present to varying degrees across glioma subtypes—may introduce bias into classification performance.

## VI. Conclusion

We present MOXAI, a multi-modal explainable AI framework for classifying adult-type diffuse gliomas, which integrates DNA methylation and copy number variation (CNV) data derived from DNA methylation arrays, alongside single-modality variants. Using ResNet50, DINOv2, and Graph Attention Network (GAT) models trained on both random ordered and genome-ordered probe matrices from TCGA, we demonstrate four key findings: (1) multimodal integration of CNV and methylation data improves classification stability over single-modality models; (2) DNA methylation alone is more discriminative than CNV alone for glioma subtype classification; (3) DINOv2 generalizes better than ResNet50, achieving 94.25% accuracy on independent test samples compared to 85.94%; and (4) ResNet50-based CAM and Grad-CAM outputs are more interpretable than those from DINOv2, aligning more closely with established molecular features of glioma subtypes. Together, these results position MOXAI as a systematic, clinically relevant approach to multi-omics integration with built-in interpretability, laying the groundwork for next-generation molecular diagnostic models capable of within-group tumor classification and risk assessment from a single array experiment. Future work will extend this framework to explore graph network variants (GIN, SAGE, GCN) and evaluate additional interpretability approaches such as Integrated Gradients (IG) and SHAP.

## References

[1] D. N. Louis *et al.*, “The 2021 WHO classification of tumors of the central nervous system: A summary,” *Neuro-Oncology*, vol. 23, no. 8, pp. 1231–1251, 2021.

[2] D. Capper *et al.*, “DNA methylation-based classification of central nervous system tumours,” *Nature*, vol. 555, pp. 469–474, 2018.

[3] Z. Alom, Q. T. Tran, A. K. Bag, J. T. Lucas, and B. A. Orr, “Predicting methylation class from diffusely infiltrating adult gliomas using multimodality MRI data,” *Neuro-Oncology Advances*, vol. 5, p. vdad045, 2023.

[4] P. A. Toussaint, F. Leiser, S. Thiebes, M. Schlesner, B. Brors, and A. Sunyaev, “Explainable artificial intelligence for omics data: A systematic mapping study,” *Briefings in Bioinformatics*, vol. 25, no. 1, p. bbad453, 2024.

[5] J. W. Brat *et al.*, “Comprehensive, integrative genomic analysis of diffuse lower-grade gliomas,” *New England Journal of Medicine*, vol. 372, no. 26, pp. 2481–2498, 2015.

[6] E. Athieniti and G. M. Spyrou, “A guide to multi-omics data collection and integration for translational medicine,” *Computational and Structural Biotechnology Journal*, vol. 21, pp. 134–149, 2023.

[7] B. Zhou, A. Khosla, A. Lapedriza, A. Oliva, and A. Torralba, “Learning deep features for discriminative localization,” in *Proc. IEEE Conf. Comput. Vis. Pattern Recognit. (CVPR)*, 2016, pp. 2921–2929.

[8] M. P. Mariani, J. A. Chen, Z. Zhang, S. C. Pike, and L. A. Salas, “MethylMasteR: A comparison and customization of methylation-based copy number variation calling software in cancers harboring large-scale chromosomal deletions,” *Frontiers in Bioinformatics*, vol. 2, p. 859828, 2022.

[9] K. He, X. Zhang, S. Ren, and J. Sun, “Deep residual learning for image recognition,” in *Proc. IEEE Conf. Comput. Vis. Pattern Recognit. (CVPR)*, 2016, pp. 770–778.

[10] D. T. W. Jones *et al.*, “Comprehensive analysis of paediatric low-grade glioma defines disparate SHH-, FGFR-, and BRAF-driven molecular subgroups,” *Nature Communications*, vol. 14, p. 4954, 2023.

[11] M. Oquab, T. Darcet, T. Moutakanni, H. Vo, M. Szafraniec, V. Khalidov, P. Fernandez, *et al.*, “DINOv2: Learning robust visual features without supervision,” *arXiv preprint arXiv:2304.07193*, 2023.

[12] D. P. Kingma and J. Ba, “Adam: A method for stochastic optimization,” *arXiv preprint arXiv:1412.6980*, 2014.

[13] D. Shen, G. Wu, and H.-I. Suk, “Deep learning in medical image analysis,” *Annual Review of Biomedical Engineering*, vol. 19, pp. 221–248, 2017.

[14] D. Acharya and A. Mukhopadhyay, “A comprehensive review of machine learning techniques for multi-omics data integration: Challenges and applications in precision oncology,” *Briefings in Functional Genomics*, vol. 23, no. 5, pp. 549–560, 2024.

[15] R. R. Selvaraju, A. Das, R. Vedantam, M. Cogswell, D. Parikh, and D. Batra, “Grad-CAM: Why did you say that?,” *arXiv preprint arXiv:1611.07450*, 2016.

[16] Q. T. Tran, A. Breuer, T. Lin, R. Tatevossian, S. J. Allen, M. Clay, *et al.*, “Comparison of DNA methylation based classification models for precision diagnostics of central nervous system tumors,” *npj Precision Oncology*, vol. 8, no. 1, p. 218, 2024.

[17] J. G. Cairncross, K. Ueki, M. C. Zlatescu, D. K. Lisle, D. M. Finkelstein, R. R. Hammond, J. S. Silver, *et al.*, “Specific genetic predictors of chemotherapeutic response and survival in patients with anaplastic oligodendrogliomas,” *J. Nat. Cancer Inst.*, vol. 90, no. 19, pp. 1473–1479, 1998.

[18] P. Veličković, G. Cucurull, A. Casanova, A. Romero, P. Lio, and Y. Bengio, “Graph attention networks,” *arXiv preprint arXiv:1710.10903*, 2017.

[19] A. Eckhardt, R. Drexler, M. Schoof, N. Struve, D. Capper, C. Jelgersma, *et al.*, “Mean global DNA methylation serves as independent prognostic marker in IDH-wildtype glioblastoma,” *Neuro-Oncology*, vol. 26, no. 3, pp. 503–513, 2024.